\documentclass{article}
\usepackage{ijcai25}

\usepackage{times}
\usepackage{soul}
\usepackage{url}
\usepackage[hidelinks]{hyperref}
\usepackage[utf8]{inputenc}
\usepackage[small]{caption}
\usepackage{graphicx}
\usepackage{amsmath}
\usepackage{amsthm}
\usepackage{booktabs}
\usepackage{algorithm}
\usepackage{algorithmic}
\usepackage{amsmath, amssymb}
\usepackage[table]{xcolor}
\usepackage[switch]{lineno}

\title{SketchAgent: Generating Structured Diagrams from Hand-Drawn Sketches}

\author{
    Cheng Tan$^{1,2}$\thanks{Equal contribution}, Qi Chen$^{3*}$, Jingxuan Wei$^{3*\dagger}$, Gaowei Wu$^{3*}$, Zhangyang Gao$^{1,2}$, Siyuan Li$^{1,2}$, Bihui Yu$^{3}$, Ruifeng Guo$^{3}$, Stan Z. Li$^{1}$\thanks{Corresponding author.}
    \affiliations
    $^1$Westlake University, $^2$Zhejiang University, $^3$University of Chinese Academy of Sciences
    \emails
    tancheng@westlake.edu.cn, weijingxuan20@mails.ucas.edu.cn
}

\begin{document}

\maketitle

\begin{abstract}
Hand-drawn sketches are a natural and efficient medium for capturing and conveying ideas. Despite significant advancements in controllable natural image generation, translating freehand sketches into structured, machine-readable diagrams remains a labor-intensive and predominantly manual task. The primary challenge stems from the inherent ambiguity of sketches, which lack the structural constraints and semantic precision required for automated diagram generation. To address this challenge, we introduce SketchAgent, a multi-agent system designed to automate the transformation of hand-drawn sketches into structured diagrams. SketchAgent integrates sketch recognition, symbolic reasoning, and iterative validation to produce semantically coherent and structurally accurate diagrams, significantly reducing the need for manual effort. To evaluate the effectiveness of our approach, we propose the Sketch2Diagram Benchmark, a comprehensive dataset and evaluation framework encompassing eight diverse diagram categories, such as flowcharts, directed graphs, and model architectures. The dataset comprises over 6,000 high-quality examples with token-level annotations, standardized preprocessing, and rigorous quality control. By streamlining the diagram generation process, SketchAgent holds great promise for applications in design, education, and engineering, while offering a significant step toward bridging the gap between intuitive sketching and machine-readable diagram generation. The benchmark is released at \href{https://huggingface.co/datasets/DiagramAgent/Sketch2Diagram-Benchmark}{this link}.
\end{abstract}
\vspace{-4mm}

\section{Introduction}

Hand-drawn sketches are a natural and powerful medium for rapidly conveying ideas, serving as a universal language in creative, technical, and educational workflows~\cite{xu2022deep,zhao2022analysis,zhao2024huber,tan2024peer}. From rough brainstorming sessions to preliminary engineering designs, sketches offer an intuitive way to externalize concepts. However, translating these informal and ambiguous drawings into structured, machine-readable diagrams remains an open challenge. Unlike natural image generation tasks~\cite{cao2024controllable,huang2024controllable,li2019controllable}, which have seen remarkable progress in recent years through techniques such as controllable natural image generation, the sketch-to-diagram task demands more than just visual fidelity—it requires understanding and formalizing the underlying structural and semantic relationships inherent to diagrams.

\begin{figure*}[t]
    \centering
    \includegraphics[width=0.96\linewidth]{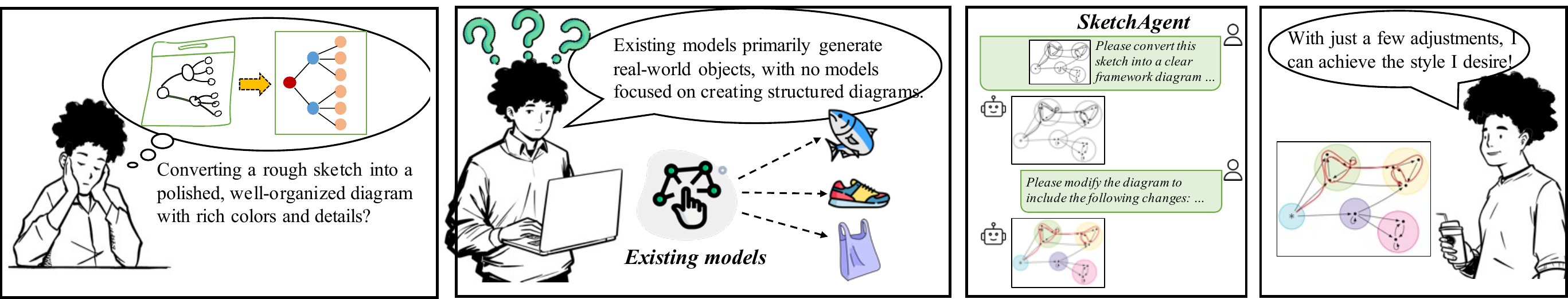}
    \vspace{-2mm}
    \caption{SketchAgent automates the transformation of hand-drawn sketches into structured diagrams.}
    \vspace{-4mm}
    \label{fig:intro}
\end{figure*}

We introduce a new task, sketch-to-diagram generation, which involves converting a hand-drawn sketch into a structured, machine-readable diagram. As shown in Figure~\ref{fig:intro}, this task differs fundamentally from controllable natural image generation, as it focuses not on generating aesthetically pleasing visuals but on synthesizing a precise, semantically meaningful diagram that adheres to specific structural rules. This transformation requires solving several core challenges: \textbf{(1) handling the inherent ambiguity and variability in freehand sketches, (2) preserving the spatial and structural relationships between diagram components, and (3) producing an output that is both syntactically valid and semantically aligned with the user's intent}. These challenges make sketch-to-diagram generation a highly specialized and underexplored problem, distinct from existing works.

To address the lack of standardized resources for sketch-to-diagram research, we introduce the Sketch2Diagram Benchmark, a comprehensive dataset and evaluation framework designed to support the development and assessment of models for this task. The dataset spans eight diverse diagram categories, including flowcharts, directed graphs, and model architectures, and consists of over 6,000 high-quality examples. Each example includes a hand-drawn sketch paired with its corresponding structured diagram representation. The dataset is meticulously curated, featuring token-level annotations, standardized preprocessing, and rigorous quality control, ensuring its reliability for both training and evaluation purposes.

Building on this benchmark dataset, we propose SketchAgent, an end-to-end system for automating the transformation of hand-drawn sketches into structured diagrams. The system begins by converting an input sketch into a symbolic code representation, which abstracts its structural and spatial properties into a machine-readable format, bridging the gap between informal freehand drawings and precise computational diagrams. From there, SketchAgent performs iterative refinement to improve the accuracy, coherence, and validity of the code representation, ensuring the final diagram adheres to the user's intent while satisfying all structural constraints. 

Our main contributions are as follows:
\begin{itemize}
    \item We formally define the task of converting hand-drawn sketches into structured diagrams, distinguishing it from related tasks such as controllable image generation.
    \item We introduce a benchmark dataset of hand-drawn sketches and their corresponding structured diagram representations, offering a standardized resource for training and evaluation.
    \item We propose SketchAgent, a modular system that automates the sketch-to-diagram transformation process, integrating sketch recognition, symbolic reasoning, iterative refinement, and verification into a unified pipeline.
\end{itemize}

The proposed sketch-to-diagram framework has wide-ranging applications in fields such as engineering, architecture, education, and design, where structured diagrams are integral to workflows. 

\section{Related Work}

\subsection{Controllable Image Generation}
Controllable image generation aims to synthesize images that adhere to specific constraints, such as object attributes, spatial arrangements, and style consistency~\cite{cao2024controllable,huang2024controllable}. Existing methods can be categorized into three main approaches: GAN-based, diffusion-based, and multi-modal fusion techniques. Early GAN-based methods, such as ControlGAN~\cite{li2019controllable}, introduced fine-grained text-conditioned image synthesis but suffered from instability and mode collapse. Diffusion models have since become the dominant paradigm, offering more stable and high-quality generation. Diffusion Self-Guidance~\cite{epstein2023diffusion} and MultiDiffusion~\cite{bar2023multidiffusion} enable explicit control over object positioning and spatial structure, while Control-GPT~\cite{zhang2023controllable} leverages GPT-4-generated sketches for improved spatial consistency. Other approaches integrate LLMs or additional modalities for enhanced control, such as MoMA~\cite{song2025moma}, which fuses textual and visual embeddings, and MM-Diff~\cite{wei2024mm}, which refines personalization through CLIP-based representations. Furthermore, PALP~\cite{arar2024palp} enhances alignment with complex textual prompts by optimizing cross-modal score matching. Despite these advancements, existing approaches primarily focus on photorealistic image synthesis, making them insufficient for structured and logic-constrained generation tasks like diagrams~\cite{cao2024controllable,huang2024controllable}. While diffusion-based models offer control over spatial attributes~\cite{epstein2023diffusion,bar2023multidiffusion}, they lack explicit structural reasoning capabilities required for diagram generation.

\subsection{Controllable Code Generation}
Controllable code generation aims to produce structured and executable code while adhering to specific constraints, such as syntax correctness, structural consistency, and execution validity~\cite{shin2021survey,wei2024words}. Language model-based approaches leverage large-scale pre-trained models to improve code synthesis. Magicoder~\cite{wei2024magicoder} enhances Python and multi-language code generation through OSS-INSTRUCT, while VeriGen~\cite{thakur2024verigen} tailors large language models for Verilog synthesis by curating specialized training datasets. Structure-aware methods further refine code generation by integrating abstract syntax trees and data flow graphs. StructCoder~\cite{tipirneni2024structcoder} introduces a structure-aware self-attention mechanism, and CoTexT~\cite{phan2021cotext} applies multi-task learning to enhance text-to-code understanding. Planning-based techniques decompose complex tasks into stepwise solutions, as seen in Self-Planning Code Generation~\cite{jiang2024self}, while reinforcement learning-based approaches such as CodeRL~\cite{le2022coderl} optimize model adaptation through reward-based fine-tuning. Execution-enhanced methods ensure generated code correctness by leveraging runtime validation. MBR-EXEC~\cite{shi2022natural} employs minimum Bayesian risk decoding based on execution results, whereas CODET~\cite{chen2022codet} generates test cases to filter invalid code. Besides, real-world integration studies, such as In-IDE Code Generation~\cite{xu2022ide}, evaluate the practical utility of code generation models.

While these methods advance code generation in terms of syntax, semantics, and execution fidelity, they remain constrained to text-based inputs, lacking the capability to synthesize code from sketch-based conceptualizations. Furthermore, existing controllable code generation approaches do not inherently support structured diagram generation, limiting their applicability in domains requiring logical and hierarchical visual representations. \cite{ghosh2018automated,almazroi2021class} have focused on extracting structured representations from textual descriptions, but these methods do not generalize to sketch-driven workflows. In contrast, our work introduces the first sketch-to-diagram pipeline, leveraging an intermediate code representation to ensure structural coherence and semantic fidelity in diagram synthesis.

\begin{figure*}[ht]
    \centering
    \includegraphics[width=\linewidth]{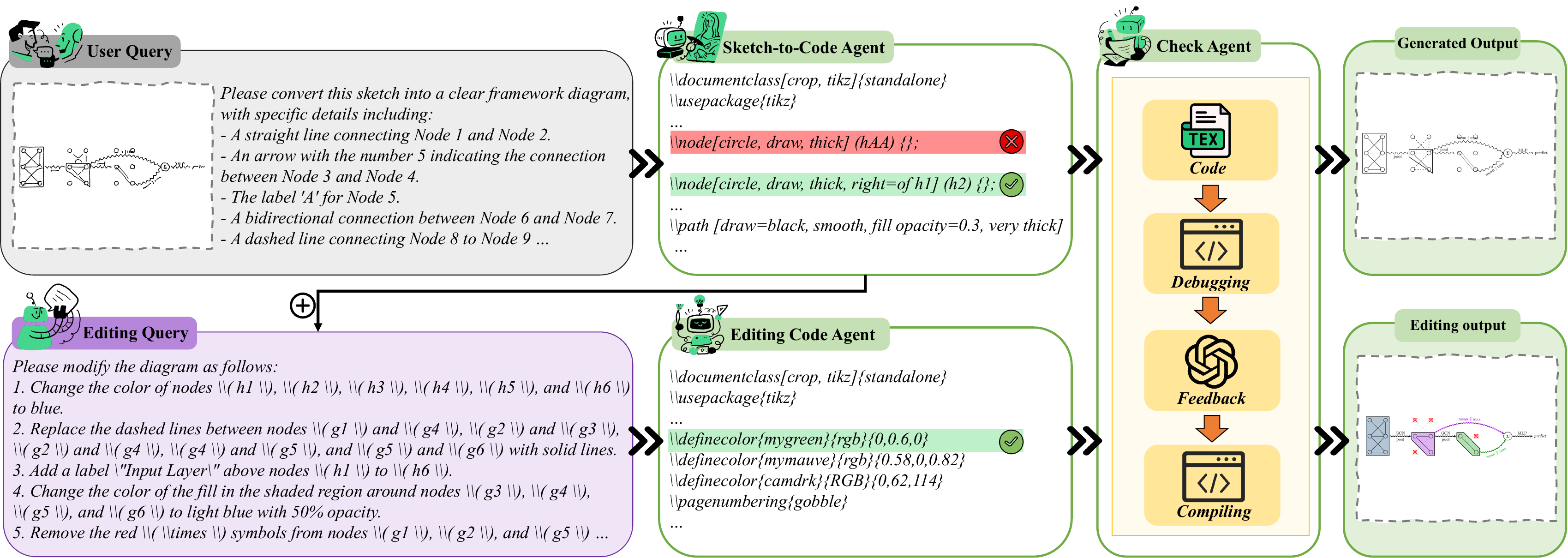}
    \caption{The SketchAgent pipeline, consisting of three main modules: Sketch-to-Code Agent, Editing Code Agent, and Check Agent.}
    \label{fig:pipeline}
\end{figure*}

\section{Method}

SketchAgent is designed to automate the transformation of hand-drawn sketches into structured diagrams through a multi-stage process. The system consists of three primary modules: the Sketch-to-Code Agent, the Editing Code Agent, and the Check Agent, each responsible for specific tasks in the pipeline. Given a sketch \( S \) and a user-specified instruction set \( Q \), SketchAgent generates an initial code representation, refines it based on additional instructions, and verifies the final output before rendering the structured diagram. The overall workflow is illustrated in Figure~\ref{fig:pipeline}.




\subsection{Sketch-to-Code Agent}

The Sketch-to-Code Agent maps a hand-drawn sketch \( S \) and an instruction set \( Q \) to an initial code representation \( C_k \), capturing the structural semantics of the sketch. This process is formulated as:
\begin{equation}
    C_k = \mathcal{F}_k(S, Q),
\end{equation}
where \( \mathcal{F}_k \) represents the transformation function. The output \( C_k \) is modeled as a sequence of tokens, where each token corresponds to a diagram component or an attribute.

To ensure the generated code aligns with the expected structure, we define the objective as minimizing the negative log-likelihood of the sequence:
\begin{equation}
    \mathcal{L}_k = - \mathbb{E}_{C_k \sim P(C \mid S, Q)} \sum_{t=1}^{T} \log P(C_k^{(t)} \mid C_k^{(<t)}, S, Q),
\end{equation}
where \( T \) is the sequence length, \( C_k^{(t)} \) is the \( t \)-th token, and \( P(\cdot) \) denotes its conditional probability given the preceding sequence. Optimizing this objective ensures \( C_k \) remains structurally valid and semantically aligned with \( S \).

\subsection{Editing Code Agent}

The Editing Code Agent refines the initial code representation \( C_k \) based on an additional instruction set \( Q' \), generating an updated version \( C_e \). This process is formalized as:
\begin{equation}
    C_e = \mathcal{F}_e(C_k, Q'),
\end{equation}
where \( \mathcal{F}_e \) represents the refinement function. The objective is to minimize the discrepancy between \( C_e \) and the expected output, ensuring modifications align with the intended diagram structure. To achieve this, we minimize the negative log-likelihood of the sequence:
\begin{equation}
    \mathcal{L}_e = - \sum_{t=1}^{T} \log P(C_e^{(t)} \mid C_e^{(<t)}, C_k, Q'),
\end{equation}
where \( T \) is the sequence length, \( C_e^{(t)} \) is the \( t \)-th token, and \( P(\cdot) \) denotes its conditional probability given the preceding sequence and input conditions. Optimizing this objective ensures \( C_e \) effectively integrates the modifications specified in \( Q' \) while preserving the structural integrity of \( C_k \).

\subsection{Check Agent}

The Check Agent verifies and refines the generated code \( C_e \) to produce the final executable representation \( C_f \). This process is formalized as:

\begin{equation}
    C_f = \mathcal{F}_f(C_e),
\end{equation}
where \( \mathcal{F}_f \) denotes the verification and debugging function. The goal is to ensure \( C_f \) is syntactically correct, executable, and aligned with the original sketch \( S \) and instructions \( Q \).

To achieve this, the Check Agent begins by performing syntax validation and debugging to ensure that the generated code representation \( C_e \) is syntactically correct and compilable. If \( C_e \) fails to compile, it is sent back for regeneration by either the Sketch-to-Code Agent or the Editing Code Agent. Once compilation succeeds, we use GPT-4o to compare the compiled diagram \( D \) with the original sketch \( S \) and the user instructions \( Q \). If the generated diagram aligns with the expected structure, the process is finalized; otherwise, the Check Agent triggers a fallback mechanism, prompting the responsible agent to regenerate \( C_e \). This iterative validation process guarantees that only structurally coherent and semantically correct diagrams are finalized.

\begin{figure*}[t]
    \centering
    \includegraphics[width=\textwidth]{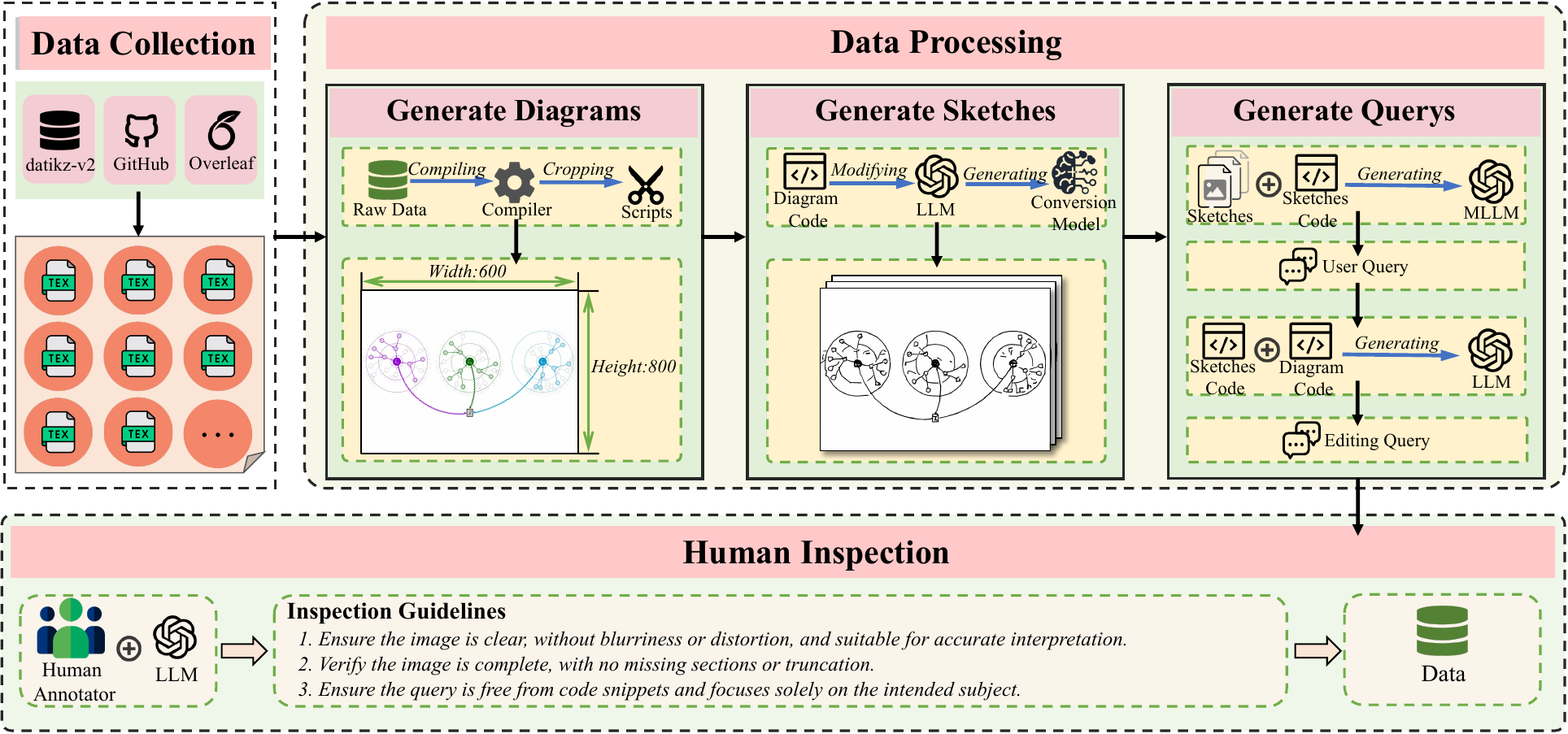}
    \caption{The data collection and processing pipeline for the Sketch2Diagram Benchmark.}
    \label{fig:data_pipeline}
\end{figure*}

\section{Sketch2Diagram Benchmark}

The Sketch2Diagram Benchmark introduces a comprehensive dataset and evaluation framework for sketch-to-diagram tasks. It features eight diverse diagram categories, standardized with rigorous quality control and token-level statistics. Clear metrics evaluate sketch-to-code and code editing tasks, ensuring thorough assessment. The data collection and processing pipeline for the Sketch2Diagram Benchmark is meticulously designed to ensure a comprehensive and high-quality dataset. This process is divided into three key stages: \textit{Data Collection}, \textit{Data Processing}, and \textit{Human Inspection}, as illustrated in Figure~\ref{fig:data_pipeline}. 

\paragraph{Data Collection.} 
The first stage involves gathering open-source \texttt{.tex} files of logical diagrams from multiple repositories, including datikz-v2, GitHub, and Overleaf. These sources provide a diverse range of diagrams across various domains, ensuring the dataset's broad applicability. The collected \texttt{.tex} files are then compiled into diagram images using standard LaTeX compilers. This step guarantees that the diagrams accurately reflect their logical structures as intended by their original creators.

\begin{figure}[!ht]
    \centering
    \includegraphics[width=\linewidth]{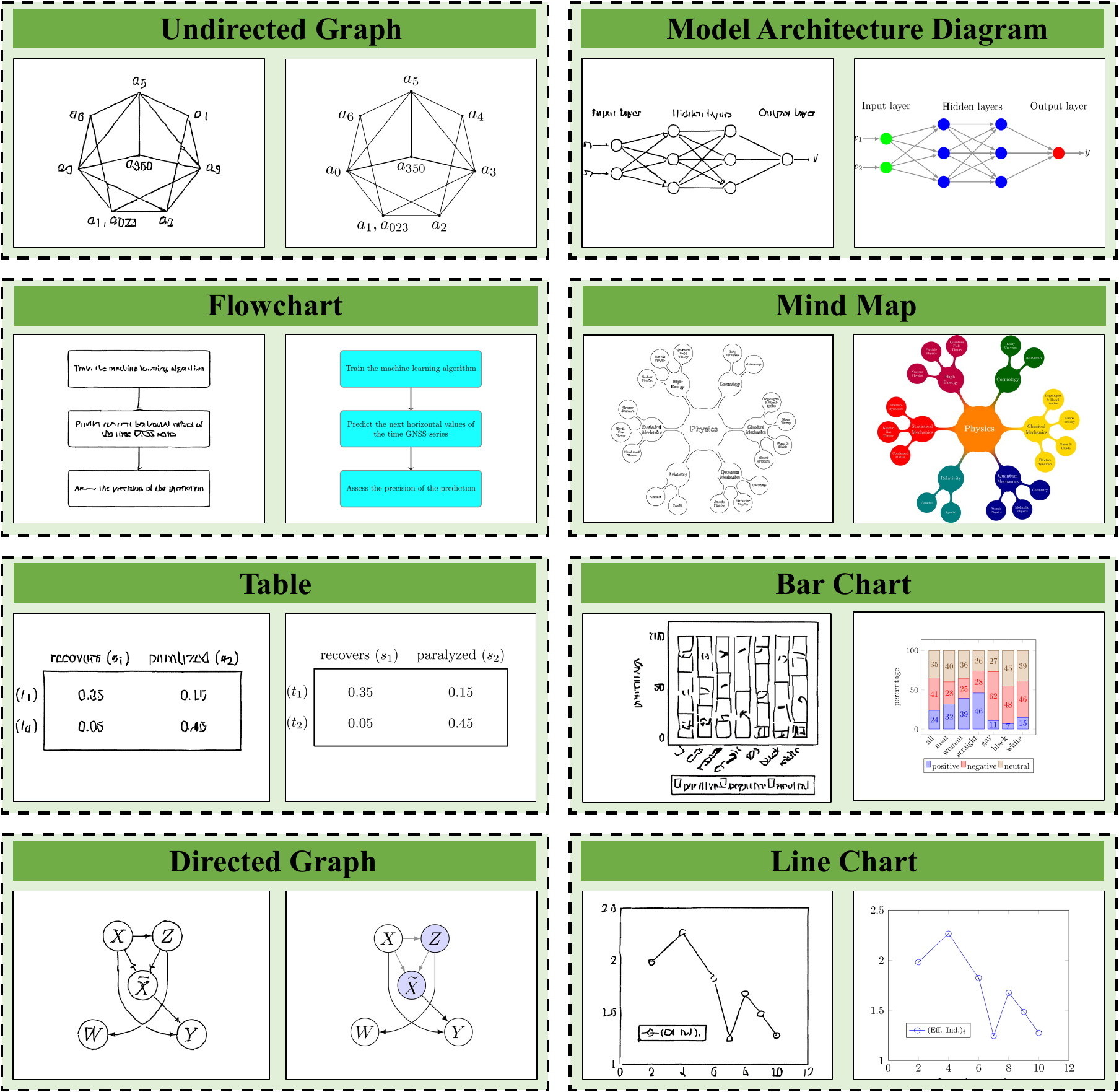}
    \caption{Examples of diagram types in the Sketch2Diagram.}
    \label{fig:sample_types}
\end{figure}

\paragraph{Data Processing.} 
We standardize the compiled diagrams to ensure uniformity and facilitate sketch-to-diagram tasks. Images are cropped to remove blank spaces and resized to 800$\times$600 pixels. Colors and intricate details are stripped from diagrams to produce simplified sketch representations. GPT-4O generates corresponding sketch codes, ensuring consistency with the original diagrams. For model evaluation, we create two query types: 
(1) \textbf{User Queries}, which pair sketch images with codes to add supplementary details; and 
(2) \textbf{Editing Queries}, which capture differences between sketch and original diagram codes to guide refinement.

\paragraph{Human Inspection.}
The final stage involves a rigorous manual inspection process to ensure data quality. Human annotators adhere to three strict guidelines: 
First, images must be clear, free of blurriness or distortion, to ensure accurate interpretation. Second, images must be complete, with no missing sections or truncations. Third, queries are reviewed to exclude code snippets and focus on descriptive elements relevant to the task. These measures ensure the dataset's reliability and suitability for  model evaluation.

\subsection{Data Analysis}

\begin{figure}[ht]
    \centering
    \vspace{-2mm}
    \includegraphics[width=\linewidth]{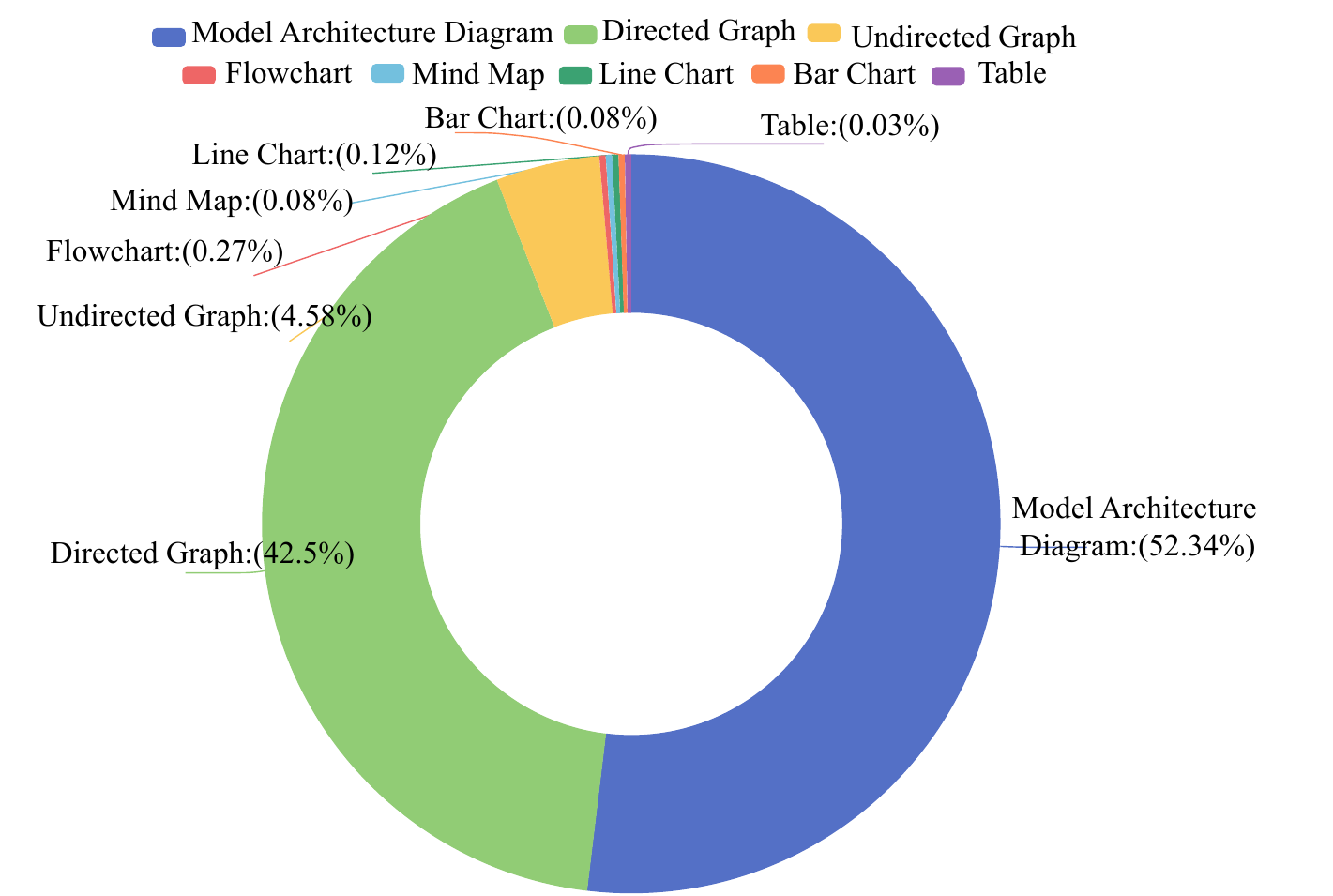}
    \caption{Category distribution in Sketch2Diagram.}
    \vspace{-3mm}
    \label{fig:category_proportions}
\end{figure}

\paragraph{Diversity and Imbalance Challenges.} 
Figures~\ref{fig:sample_types} and~\ref{fig:category_proportions} illustrate the dataset's diversity and distribution. Figure~\ref{fig:sample_types} showcases examples from eight diagram categories, including undirected graphs, model architecture diagrams, and flowcharts, highlighting their real-world relevance. Figure~\ref{fig:category_proportions} reveals an imbalance, with model architecture diagrams (52.34\%) and directed graphs (42.5\%) dominating, while categories like flowcharts and mind maps are underrepresented.

\paragraph{Token Length Statistics.} 

Table~\ref{tab:token_statistics} summarizes token length statistics for the Sketch2Diagram dataset, categorized by sketch-to-code (S2C) and code-editing (C2C) tasks. The dataset contains a total of 4824 training samples and 1206 test samples. Query lengths range from a minimum of 28 tokens for S2C tasks to a maximum of 170,894 tokens for C2C tasks. Answer lengths exhibit similar variability, with maximum values reaching 170,371 tokens. These diverse token lengths reflect the dataset's ability to evaluate models across tasks with varying complexities and input-output structures.

\begin{table}[ht]
    \centering
    \vspace{-1mm}
    \caption{Key statistics of the Sketch2Diagram dataset.}
    \vspace{-1mm}
    \label{tab:token_statistics}
    \resizebox{\linewidth}{!}{
    \begin{tabular}{lcccc}
        \toprule
        \textbf{Statistic} & \textbf{Train S2C} & \textbf{Train C2C} & \textbf{Test S2C} & \textbf{Test C2C} \\
        \midrule
        \multicolumn{5}{l}{\textbf{Total Samples}} \\
        Sample Count & 4824 & 4824 & 1206 & 1206 \\
        \midrule
        \multicolumn{5}{l}{\textbf{Query Length (tokens)}} \\
        Minimum & 28 & 165 & 39 & 208 \\
        Maximum & 1118 & 170894 & 1115 & 28715 \\
        Average & 243.9 & 1161.0 & 243.2 & 1074.3 \\
        \midrule
        \multicolumn{5}{l}{\textbf{Answer Length (tokens)}} \\
        Minimum & 119 & 125 & 133 & 133 \\
        Maximum & 170371 & 170371 & 28633 & 28633 \\
        Average & 1010.0 & 1064.1 & 922.4 & 977.8 \\
        \bottomrule
    \end{tabular}}
    \vspace{-2mm}
\end{table}

\subsection{Evaluation Metrics}
We employ a comprehensive set of evaluation metrics. For sketch-to-code, Pass@1 measures functional correctness, while ROUGE-L, BLEU, CodeBLEU (C-BLEU), chrF, Edit Distance (ED), BLEURT, and RUBY evaluate textual and semantic similarity. For code editing, we extend these metrics to include diagram fidelity measures such as FID, KID, CLIP-FID (C-FID), Inception Score (IS), LPIPS, and SSIM. 

\section{Experiment}

\paragraph{Setup} The Sketch-to-Code Agent is based on Qwen2-VL-7B~\cite{qwen2-vl}, while the Editing Code Agent utilizes Qwen2.5-Coder-7B~\cite{qwen25}. Both agents were fine-tuned over four epochs on a 4$\times$80GB A100 GPU setup. The input token length for both agents is set to 4096 tokens.

\paragraph{Model} For sketch generation, the Sketch-to-Code Agent is compared against several state-of-the-art models, including Yi-VL~\cite{yi-vl-34b}, Qwen2-VL~\cite{qwen2-vl}, Internlm-Xcomposer2.5~\cite{internlm}, Llama-3.2-Vision~\cite{llama3.2-vision-instruct}, Phi-3.5-Vision~\cite{phi}, Llava-v1.6~\cite{llava-v1.6-34b}, Cogvlm2-Llama3~\cite{cogvlm2}, and DeepSeek-VL~\cite{deepseek-vl-7b-chat}, with close-source models such as GPT-4o~\cite{gpt4o}, GLM-4-plus~\cite{glm-4-plus}, and Gemini-1.5-Pro~\cite{gemini} also implemented. For the sketch editing task, the Editing Code Agent is assessed alongside specialized code models, including Qwen2.5-Coder~\cite{qwen25}, DeepSeek-Coder-Instruct~\cite{deepseekcoder}, Code-Llama~\cite{codellama}, WizardCoder~\cite{wizardcoder}, CodeGeeX4-All~\cite{codegeex4}, Starcoder2~\cite{starcoder}, Yi-Coder~\cite{yi-coder}, Llama 3.1~\cite{llama3_1}, Baichuan2~\cite{baichuan}, Internlm2.5~\cite{internlm2_5}, Yi-1.5~\cite{yi-1.5}, and Qwen2~\cite{qwen2}, and close-source models like GPT-4o~\cite{gpt4o}, DeepSeekV2.5~\cite{deepseek-v2.5}, GLM-4-Plus ~\cite{glm-4-plus}, and Gemini-1.5-Pro~\cite{gemini}. 

\begin{table*}[ht]
\caption{(a) Main results: Performance comparison on sketch generation with open-source and closed-source models on key code generation metrics. The best result is highlighted in bold. (b) Ablation study: Performance under different component configurations.}
\label{image2code}
\small
{\renewcommand\baselinestretch{1.}\selectfont
\setlength{\tabcolsep}{2.2mm}{
\begin{tabular}{lccccccccc}
\toprule
Model &
  \multicolumn{1}{l}{Size} &
  \multicolumn{1}{l}{Pass@1↑} &
  \multicolumn{1}{l}{ROUGE-L↑} &
  \multicolumn{1}{l}{C-BLEU↑} &
  \multicolumn{1}{l}{BLEU↑} &
  \multicolumn{1}{l}{ED↓} &
  \multicolumn{1}{l}{chrF↑} &
  \multicolumn{1}{l}{BLEURT↑} &
  \multicolumn{1}{l}{RUBY↑} \\ \midrule
\multicolumn{10}{c}{(a) Main results}                                       \\ \midrule
Yi-VL-34B                     & 34B & 0.25  & 33.68 & 77.79 & 8.23  & 94.17 & 19.87 & 37.28 & 21.52 \\
Qwen2-VL-7B-Instruct          & 7B  & 52.40 & 33.72 & 75.19 & 7.37  & 90.38 & 26.26 & 32.91 & 21.97 \\
internlm-xcomposer2d5-7b      & 7B  & 0.08  & 30.32 & 78.65 & 3.61  & 94.50 & 19.98 & 37.18 & 18.57 \\
Llama-3.2-11B-Vision-Instruct & 11B & 40.09 & 39.17 & 80.36 & 16.61 & 87.35 & 28.09 & 37.64 & 26.05 \\
Phi-3.5-vision-instruct       & 4B  & 16.64 & 13.41 & 68.95 & 5.20  & 97.37 & 12.65 & 34.27 & 7.82  \\
llava-v1.6-34b                & 34B & 8.51  & 24.63 & 76.75 & 10.68 & 96.73 & 27.49 & 33.75 & 13.96 \\
cogvlm2-llama3-chat-19B       & 19B & 0.00  & 12.60 & 70.57 & 3.16  & 98.03 & 12.27 & 32.73 & 7.35  \\
deepseek-vl-7b-chat           & 7B  & 0.33  & 26.94 & 80.74 & 12.06 & 95.68 & 22.47 & 37.18 & 15.21 \\
GPT-4o                        & -   & 51.12 & 34.42 & 79.56 & 12.65 & 92.20 & 29.51 & 33.40 & 20.85 \\
Gemini-1.5-pro                & -   & 61.94 & 39.80 & 81.02 & 12.64 & 89.10 & 30.61 & 32.53 & 25.86 \\
GLM-4V-Plus                   & -   & 66.09 & 40.53 & 81.30 & 12.73 & 88.16 & 30.01 & 33.00 & 26.70 \\
\rowcolor[HTML]{E7ECE4}  \textbf{SketchAgent} &
  7B &
  {\textbf{82.34}} &
  {\textbf{52.96}} &
  {\textbf{86.02}} &
  {\textbf{29.61}} &
  {\textbf{73.16}} &
  {\textbf{47.88}} &
  {\textbf{46.34}} &
  {\textbf{41.17}} \\ \midrule
\multicolumn{10}{c}{(b) Ablation study}                                     \\ \midrule
\rowcolor[HTML]{F0F4EE} w/o GPT-4o                  & 7B  & 81.18 & 52.90 & 86.01 & 29.34 & 73.16 & 47.36 & 46.28 & 41.05 \\
\rowcolor[HTML]{F0F4EE} w/o Compiler                & 7B  & 80.85 & 52.83 & 86.01 & 29.33 & 73.23 & 47.78 & 46.25 & 41.02 \\
\rowcolor[HTML]{F0F4EE} w/o GPT-4o \& Compiler      & 7B  & 78.52 & 52.39 & 85.98 & 25.88 & 73.36 & 47.82 & 46.38 & 40.78 \\ \bottomrule
\end{tabular}}
}
\end{table*}

\subsection{Sketch generation}

\paragraph{Main Results} SketchAgent leverages compiler principles and integrates GPT-4o as a feedback mechanism to achieve state-of-the-art performance in translating logical structure diagrams into executable code. As demonstrated in Table~\ref{image2code}, SketchAgent significantly outperforms both open-source and close-source models across key code generation metrics. Notably, it achieves a Pass@1 score of 82.34, substantially surpassing powerful large language models such as GPT-4o (51.12), Gemini-1.5-pro (61.94), and GLM-4V-Plus (66.09). In addition, SketchAgent excels in other critical metrics, including CodeBLEU (86.02), BLEU (29.61), and BLEURT (46.34), indicating its strong ability to generate syntactically and semantically accurate code. These evaluation results underscore SketchAgent's superior capability in producing high-quality, executable code from sketch representations.

\paragraph{Ablation Study} We conduct an ablation study as detailed in Table~\ref{image2code}. The results reveal the critical role played by both the compiler module and GPT-4o feedback mechanism. Removing the compiler results in a 1.49-point drop in Pass@1, indicating the compiler's significant contribution to precise code generation. Removing the GPT-4o feedback mechanism leads to a 1.16-point decline in Pass@1, demonstrating GPT-4o's role in validating outputs. More importantly, when both the compiler and GPT-4o are removed, the Pass@1 score experiences a substantial decrease of 3.82 points, falling to 78.52, highlighting their synergistic effect on SketchAgent's performance. These findings confirm that both components are indispensable for maximizing SketchAgent’s accuracy and effectiveness in diagram-to-code generation.

\paragraph{Human Evaluation}

\begin{figure}[ht]
    \centering
    \vspace{-2mm}
    \includegraphics[width=\linewidth]{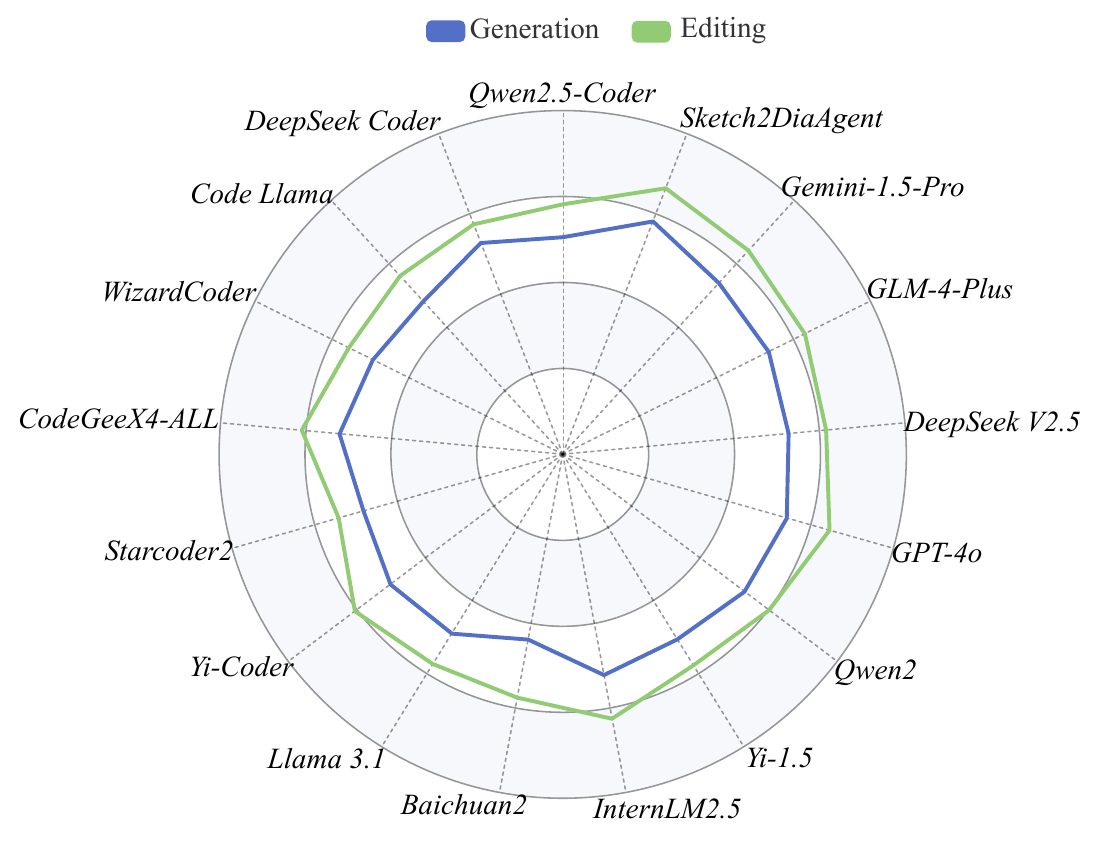}
    \caption{Human evaluation results for different models on diagram generation and Modify diagram generation tasks.}
    \label{hum}
\end{figure}

We employed three professional evaluators to assess the outputs. A score of 1 indicated the lowest quality, while 5 represented the highest quality. As shown in Figure~\ref{hum}, SketchAgent consistently achieved the highest quality based on objective assessment metrics. Furthermore, SketchAgent surpassed the accuracy of other models. Interestingly, the results revealed that SketchAgent performed significantly better on the editing task compared to the generation task. This suggests that the model excels at refining existing images to achieve the desired outcome, rather than generating entirely new objects from scratch based on instructions. This observation implies that SketchAgent iteratively optimizes outputs based on user-provided instructions, ultimately reaching the target result. It also highlights the importance of the editing task within the image generation.

\subsection{Sketch Editing}

\begin{table*}[h]
    \centering
    \vspace{-2mm}
    \caption{(a) Main results: Performance comparison on sketch editing with open-source and closed-source models on both code generation and fidelity metrics. The best result is highlighted in bold. (b) Ablation study: Performance under different component configurations.}
    \vspace{-2mm}
    \label{tab:Editing}
    \small
    {\renewcommand\baselinestretch{1.}\selectfont
    \setlength{\tabcolsep}{0.4mm}{
    \begin{tabular}{lccccccccc|ccccccc}
    \midrule
        Model & {Size} & {Pass@1↑} & {ROUGE↑} & {C-BLEU↑} & {BLEU↑} & {ED↓} & {chrF↑} & {BLEURT↑} & {RUBY↑} & {IS↑} & {FID↓} & {KID↓} & {C-FID↓} & {LPIPS↓} & {SSIM↑} \\ \midrule
        \multicolumn{16}{c}{(a) Main results}                                       \\ \midrule
        Qwen2.5-Coder & 7B & 62.19 & 89.43 & 95.59 & 78.11 & 23.26 & 88.22 & 46.25 & 82.56 & 3.82 & 246.80 & 19.59 & 64.77 & 49.68 & 0.43 \\ 
        DeepSeek Coder & 33B & 81.92 & 88.53 & 95.37 & 75.38 & 22.31 & 87.67 & 46.28 & 81.66 & 3.95 & 222.42 & 1.90 & 29.16 & 54.07 & 0.51 \\ 
        Code Llama & 34B & 59.78 & 96.23 & 96.57 & 77.20 & 6.07 & 87.76 & 73.62 & 93.66 & 3.02 & 283.48 & 2.10 & 36.63 & 44.13 & 0.28 \\ 
        WizardCoder & 15B & 92.12 & 94.30 & 96.78 & 76.39 & 10.03 & 90.85 & 73.82 & 90.84 & 4.03 & 283.48 & 2.09 & 36.63 & 44.13 & 0.28 \\ 
        CodeGeeX4-ALL & 9B & 69.32 & 86.07 & 93.24 & 63.11 & 21.85 & 84.99 & 68.64 & 79.19 & 2.70 & 209.63 & 12.45 & 53.08 & 44.29 & 2.81 \\ 
        Starcoder2 & 15B & 88.39 & 86.31 & 91.38 & 57.68 & 22.10 & 83.08 & 69.13 & 83.86 & 3.75 & 283.48 & 2.10 & 36.63 & 44.13 &  0.28 \\ 
        Yi-Coder & 9B & 66.67 & 34.17 & 39.34 & 8.14 & 30.87 & 33.83 & 49.12 & 31.68 & 3.35 & 228.94 & 2.09 & 30.11 & 53.44 &  0.44 \\ 
        Llama 3.1 & 8B & 59.95 & 71.71 & 85.77 & 68.47 & 35.35 & 69.72 & 60.95 & 65.53 & 3.68 & 238.25 & 1.37 & 26.88 & 55.61 & 0.42 \\ 
        Baichuan2 & 13B & 18.57 & 41.88 & 75.37 & 8.51 & 65.03 & 40.01 & 47.83 & 36.16 & 4.43 & 148.70 & 9.73 & 33.80 & 75.84 & 6.42 \\ 
        InternLM2.5 & 20B &  80.43 & 90.09 & 94.81 & 74.78 & 16.03 & 88.64 & 70.86 & 84.59 & 3.09 & 220.07 & 1.00 & 29.11 & 48.84 & 0.51 \\ 
        Yi-1.5 & 34B & 53.23 & 83.14 & 94.19 & 60.51 & 28.24 & 81.85 & 66.41 & 75.69 & 3.50 & 226.15 & 1.77 & 30.12 & 49.97 & 10.43 \\ 
        Qwen2 & 7B & 58.96 & 78.93 & 93.01 & 60.22 & 37.30 & 76.56 & 61.60 & 69.73 & 3.22 & 237.49 & 1.34 & 27.85 & 50.50 & 0.42 \\ 
        GPT-4o & - & 91.79 & 96.81 & 96.78 & 86.40 & 14.23 & 92.69 & 69.46 & 94.87 & 5.51 & 139.56 & 0.23 & 13.99 & 46.94 & 28.41 \\ 
        DeepSeek V2.5 & - & 83.67 & 94.30 & 96.77 & 80.54 & 17.02 & 90.85 & 71.01 & 90.84 & 4.57 & 203.27 & 0.98 & 33.18 & 42.16 &  0.43 \\ 
        GLM-4-Plus & - & 87.06 & 95.88 & 96.78 & 86.06 & 10.02 & 91.02 & 73.27 & 88.81 & \textbf{5.71} & 243.45 & 1.32 & 23.44 & 47.38 & 0.61 \\ 
        Gemini-1.5-Pro & - & 83.75 & 94.26 & 95.42 & 85.72 & 19.08 & 90.75 & 69.08 & 90.77 & 3.83 & 246.68 & 19.58 & 64.70 & 49.57 & 0.26 \\ 
    \midrule
    \rowcolor[HTML]{E7ECE4}   \textbf{SketchAgent} & 7B & \textbf{93.12} & \textbf{97.68} & \textbf{98.63} & \textbf{87.38} & \textbf{5.92} & \textbf{96.38} & \textbf{74.33} & \textbf{96.26} & 5.51 & \textbf{130.15} & \textbf{0.22} & \textbf{12.24} & \textbf{42.00} & \textbf{30.13} \\
\midrule
\multicolumn{16}{c}{(b) Ablation study}                                     \\ \midrule
\rowcolor[HTML]{F0F4EE} - w/o Feedback & 7B & 94.69 & 97.65 & 98.62 & 87.38 & 14.21 & 96.32 & 71.28 & 94.21 & 5.20 & 143.54 & 0.26 & 13.54 & 47.29 & 32.47 \\ 
\rowcolor[HTML]{F0F4EE} - w/o Compilation & 7B & 91.21 & 96.26 & 98.62 & 85.12 & 5.95 & 94.77 & 73.05 & 95.68 & 5.48& 139.48 & 0.30 & 13.03 & 47.15 & 27.09  \\ 
\rowcolor[HTML]{F0F4EE} - w/o both & 7B & 88.39 & 95.39 & 96.77 & 83.56  & 10.05  & 92.68  & 70.21  & 90.11 & 4.92  & 154.23  & 0.51 & 14.25 & 50.71  & 26.33  \\
    \bottomrule
    \end{tabular}}}
    \vspace{-1mm}
\end{table*}

\paragraph{Main Results}

SketchAgent's Editing Code Agent demonstrates high performance across several key metrics, as shown in Table~\ref{tab:Editing}. It achieves the highest Pass@1 score of 93.12, surpassing other models such as Qwen2.5-Coder (62.19) and DeepSeek Coder (81.92). Moreover, SketchAgent excels in CodeBLEU and BLEU, with scores of 98.63 and 87.38, respectively, surpassing WizardCoder (96.78 and 76.39) and Code Llama (96.57 and 77.20). In terms of visual fidelity metrics, SketchAgent performs well with a FID of 130.1494, outperforming models such as DeepSeek Coder (222.42) and Qwen2.5-Coder (246.79), demonstrating its strong visual coherence and competitiveness. These results demonstrate that SketchAgent not only excels in generating precise and syntactically accurate diagram code but also produces visually coherent and semantically meaningful diagrams.

\paragraph{Ablation Study}
The ablation study in Table~\ref{tab:Editing} evaluates the impact of feedback module and compilation module on SketchAgent’s performance. The full model achieves a Pass@1 score of 93.12, slightly lower than the 94.69 of the model without feedback, suggesting that the removal of feedback marginally improves performance. However, both models outperform the one without compilation, which scores 91.21, highlighting the importance of the compilation module. For ROUGE-L, the full model leads with 97.68, followed by the model without feedback at 97.65, while the model without compilation drops to 96.26, emphasizing the contribution of the compilation module.  In FID, the full model outperforms both ablated versions with a score of 130.15, demonstrating the importance of both components in maintaining visual coherence. The model without both feedback and compilation shows higher KID and CLIP-FID scores but sacrifices performance in other key metrics, indicating the necessity of both components for optimal performance.

\subsection{Error Analysis}

\begin{figure}[!h]
    \centering
    \vspace{-2mm}
    \includegraphics[width=0.98\linewidth]{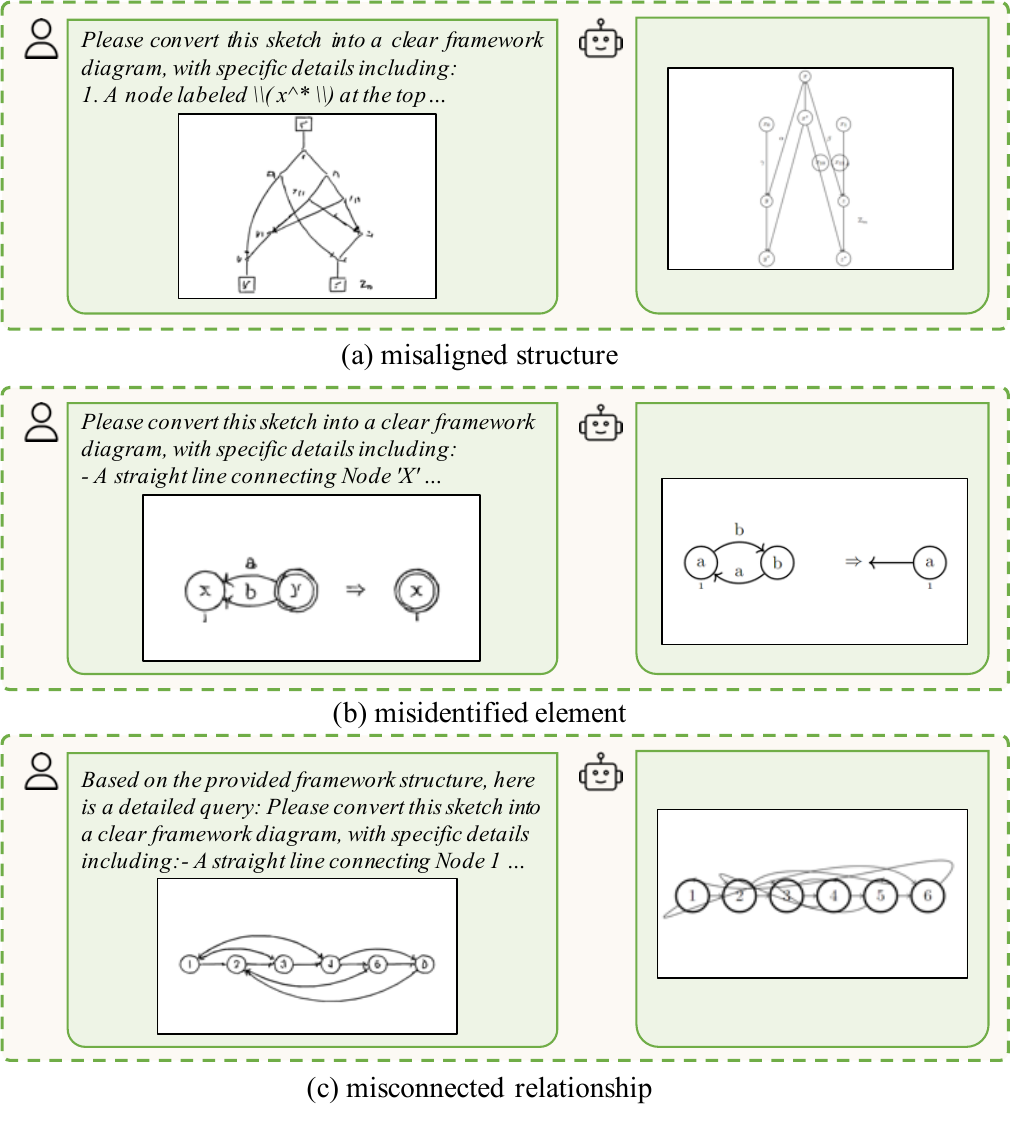}
    \vspace{-4mm}
    \caption{The error examples of SketchAgent.}
    \vspace{-4mm}
    \label{fig:error}
\end{figure}

As shown in Figure~\ref{fig:error}, SketchAgent encounters errors in three key areas: misaligned structures, misidentified elements, and misconnected relationships. Misaligned structures occur when the model fails to accurately capture the underlying structure of the objects in the sketch. This leads to incomplete or overly simplified outputs. 

Misidentified elements refer to errors in recognizing individual components of a sketch. For instance, the model may distort characters, fail to reproduce their correct proportions, or misrepresent attributes such as font style or orientation. These errors stem from the variability and ambiguity in the sketches, where subtle variations in shape can confuse recognition. Misconnected relationships arise when the system misinterprets the spatial or relational connections between elements. For example, arrows may connect incorrect components, omit intermediate nodes, or bypass key elements, leading to diagrams with logical inconsistencies. Such errors undermine the semantic integrity of the generated diagram. Due to space constraints, a more detailed error analysis is provided in the supplementary material.

\vspace{-2mm}
\section{Conclusion}

In this work, we present SketchAgent, a modular and end-to-end system for transforming hand-drawn sketches into structured, machine-readable diagrams. By addressing the core challenges of sketch-to-diagram generation, such as handling the ambiguity and variability of freehand sketches, preserving structural relationships, and ensuring semantic and syntactic validity, SketchAgent demonstrates its ability to produce accurate and semantically coherent diagrams with minimal human intervention. To evaluate the performance and capabilities of SketchAgent, we introduce the Sketch2Diagram Benchmark, a comprehensive dataset featuring over 6,000 high-quality examples spanning eight diverse diagram categories. Extensive experiments demonstrate that SketchAgent outperforms state-of-the-art models across key metrics, achieving superior accuracy and visual coherence in both sketch generation and editing tasks.

\noindent\textbf{Acknowledgements}
This work was supported by National Science and Technology Major Project (No. 2022ZD0115101), National Natural Science Foundation of China Project (No. 624B2115, No. U21A20427), Project (No. WU2022A009) from the Center of Synthetic Biology and Integrated Bioengineering of Westlake University and Integrated Bioengineering of Westlake University and Project (No. WU2023C019) from the Westlake University Industries of the Future Research Funding.

\bibliographystyle{named}
\bibliography{ijcai25}

\end{document}